\documentclass[11pt]{article}

\usepackage[]{acl}

\usepackage{times}
\usepackage{latexsym}
\usepackage{graphicx}
\usepackage{subcaption}

\usepackage[T1]{fontenc}

\usepackage[utf8]{inputenc}

\usepackage{microtype}

\usepackage{inconsolata}

\usepackage{graphicx}

\usepackage{booktabs}
\usepackage{multirow}
\usepackage{makecell}
\usepackage{array}
\usepackage{amsmath,amssymb}
\usepackage{pifont}
\usepackage{graphicx}
\usepackage{xspace}

\title{Anchorless Diversification for Parallel LLM Ideation}

\author{Fares Nabil Ibrahim, Nafis Saami Azad, Raiyan Abdul Baten$^{\dagger}$ \\
        Bellini College of Artificial Intelligence, Cybersecurity, and Computing,\\ University of South Florida, USA\\
        \texttt{\{faresibrahim, nafisazad, rbaten\}@usf.edu}\\
        \small{
   $^{\dagger}$Correspondence: \href{mailto:rbaten@usf.edu}{rbaten@usf.edu}
 }}
  
\begin{document}
\maketitle

\begin{abstract}
LLMs are increasingly used to generate candidate-idea pools for creative tasks where broad exploration is valuable. Parallel inference can be attractive in this setting when it broadens the pool while retaining quality and cost efficiency. We study inference-time controls for candidate-pool diversification, asking whether anchorless methods can rival methods that depend on observed seed ideas. Across three creative task families, we compare independent generation and semantic direction stratification with self-, peer-, and representative-anchor baselines, under neutral and population-referential divergent instructions. Population-referential divergence is a strong low-cost baseline, increasing semantic diversity while preserving quality proxies. Semantic direction stratification is stronger: a single planning call organizes generations across broad semantic directions, yielding the best diversity--quality--compute frontier. Anchored regeneration can be strong in final-pool diversity, but its advantage shrinks under full-pipeline token accounting. These results establish practical anchorless baselines for open-ended LLM ideation.
\end{abstract}

\section{Introduction}
LLMs are increasingly used for creative ideation by generating \textit{pools} of possibilities: research questions, product concepts, story premises, design directions, or campaign ideas~\citep{gottweis2026accelerating,ghareeb2026multi,suh2024luminate}. This makes parallel inference attractive: instead of scaling model parameters, a deployed system can attempt to optimally fan out many candidate generations in parallel that a human or downstream agent can compare, reject, combine, and develop~\citep{snell2025scaling,zhu2025scaling}. The value of this strategy depends on whether the pool broadens the space of possibilities without sacrificing quality or compute budget. When many candidates return to the same semantic basin, parallel generation creates an appearance of exploration while narrowing the ideas available for downstream AI creation and human--AI co-creation~\citep{doshi2024generative,anderson2024homogenization,bellemare2026divergent,padmakumar2024does,baten2026ai,ahmed2026semantic}. This failure mode is especially plausible for contemporary LLMs, whose pretraining, post-training, and preference optimization can concentrate probability mass around familiar patterns~\citep{xu2025echoes,ismayilzada2024evaluating,hou2025creativityprism,lin2025creativity,shypula2025evaluating}.

A growing literature has studied how to diversify LLM outputs in creative generation and adjacent NLP tasks. Temperature tuning is the simplest route, but often trades diversity against quality~\citep{tevet2021evaluating,chung2023increasing,zhang2021trading}. More structured methods add inference-time machinery: collective critique and voting, multi-agent debate, self-evaluation, evolutionary search, iterative prompting, or anti-anchors selected from prior outputs~\citep{lahoti2023improving,liang2023encouraging,franceschelli2024creative,bradley2024quality,chung2023increasing,hayati2024far,zhang2025noveltybench}. These methods can improve final-pool diversity, but they weaken the practical appeal of parallel inference by requiring extra calls, shared state, observed examples, or pool-level coordination. Prior work typically evaluates the returned set while leaving this hidden inference path outside the cost comparison. For scalable parallel creative ideation, the challenge is to obtain as much diversity as possible per unit of inference cost while preserving output quality.

Recent stratified-generation work offers a useful clue. In underspecified question-answering tasks, LLMs can identify meaningful partitions of a response space \emph{without first observing sample answers}, then use those partitions to improve resampling diversity and answer coverage while maintaining quality checks~\citep{wong2026simplestrat}. This suggests that a model may possess a higher-level map of plausible response regions even when its default samples concentrate in only a few of them. We adapt this approach for a broader suite of creative ideation tasks: a single planning call asks the model to propose broad semantic directions for the task, and all later candidate generations then run in parallel with the generation budget allocated evenly across those directions. The model's own semantic map therefore becomes a lightweight way to spread a candidate pool across distinct regions of a creative possibility space, without requiring seed ideas or sequential coordination.

This same intuition also motivates revisiting an intervention that prior evidence makes easy to doubt: direct instruction~\citep{guo2025benchmarking,zhang2024forcing}. Earlier prompt-only approaches often ask models to be creative or novel, but those instructions can be underspecified. Creativity may refer to the intrinsic quality of one response, whereas candidate-pool diversity is extrinsic and population-level: an idea is useful partly because it differs from what else the system would have produced~\citep{bangash2025musescorer}. Even when a model follows the instruction to move away from the default response, the instruction alone gives no explicit account of where it should go; crowding can simply reappear in the next most obvious semantic basin. We therefore use a counterfactual, population-referential instruction: generate a response that stands out from other responses that might be generated for the same task. This tests whether contemporary instruction-tuned models can use implicit knowledge of likely responses to avoid redundancy without seeing the pool they are diversifying against.

We evaluate these two anchorless controls against stronger stateful alternatives. Across alternative uses, slogans, and short stories, we generate candidate idea pools from GPT-5.4, Claude Sonnet 4.6, and Gemini 2.5 Pro. We compare independent generation, semantic direction stratification, self-anchoring, dyadic and triadic peer anti-anchoring, and shared representative anti-anchoring, each under neutral and divergent instructions. We evaluate each pool using semantic diversity metrics, task-specific quality proxies, rarefaction, and full-pipeline token accounting. This separates the final-pool question from the deployment question: a method may produce a diverse pool, yet become unattractive once planning, seed generation, and regeneration costs are counted.

Three results emerge. First, a direct divergence instruction is a strong low-cost baseline: it improves independent-pool diversity for all three models while preserving or improving automatic quality proxies. Second, semantic direction stratification is a strong anchorless diversification signal: it uses a single planning call to organize parallel generation around broad semantic directions, reaches the top tier of final-pool diversity, and yields high diversity gain per 100k pipeline tokens. Third, stratification and divergence are complementary. Their combination provides the strongest diversity--quality--compute frontier in our design and remains substantially cheaper than seed-and-regenerate pipelines. Local self- and peer-anchor methods remain powerful when seed outputs are available, but their advantage shrinks under full-pipeline token accounting. These findings establish direct divergent independent generation and equal-budget stratified divergence as practical inference-time baselines for open-ended LLM ideation.

\section{Related Work}
LLMs sample from prompt-conditioned distributions shaped by shared training data, post-training objectives, decoding parameters, and prompting protocols, often causing the model generations to collapse toward repeated themes, templates, or semantic regions~\cite{shypula2025evaluating,holtzman2019curious,kirk2023understanding,nguyen2025turning,yun2025price}. This has motivated several approaches to increasing output diversity, spanning model training and inference-time interventions.

One line of work makes diversity a model-development objective. Post-training and preference-optimization methods encourage models to prefer rare and high-quality responses, often through data selection and training objectives~\citep{wang2024weaver,chung2025modifying,lanchantin2025diverse,ismayilzada2025creative,deshpande2025diverse}. Related work on iterative self-improvement and synthetic data generation shows that repeated training on model-generated text can reduce linguistic or solution-path diversity unless diversity is built into data selection or optimization~\citep{qin2025dive,guo2024curious,guo2025benchmarking}. These approaches are important because they make diversity part of the model or data pipeline, but they are less directly available when users operate fixed, closed-weight models and must decide how to allocate inference calls at deployment time.

A second line of work increases diversity at inference time. Decoding controls such as diverse beam search, temperature, top-$k$, top-$p$, min-$p$, and adaptive temperature can increase variation, but often introduce quality, validity, or instruction-following tradeoffs~\citep{vijayakumar2016diverse,ippolito2019comparison,holtzman2019curious,tevet2021evaluating,chung2023increasing,zhang2021trading,nguyen2025turning,zhang2024edt}. Prompting and in-context approaches instead use the model's instruction-following ability to ask for novelty, avoid prior generations, critique candidate sets, incorporate negative examples, follow answer-space constraints, or search through alternative candidates~\citep{zhang2024forcing,guo2025benchmarking,lahoti2023improving,hayati2024far,zhang2024improving,ruan2025g2,suh2024luminate,wang2024guiding,bradley2024quality,franceschelli2024creative}. These methods motivate both sides of our design: direct population-referential instructions test whether models can avoid likely responses without observing a seed pool, while anchored baselines test how much is gained when prior outputs, critiques, shared state, or regeneration rounds are available. Yet this literature has rarely made the full inference path the unit of comparison: methods are often judged by the diversity of the final pool, even when that pool depends on uncounted seeding, selection, feedback, or regeneration.

A close structured predecessor is \textsc{SimpleStrat}, which shows that an LLM can identify useful partitions of a response space and use stratum-specific prompts to improve coverage in underspecified question answering~\citep{wong2026simplestrat}. We extend this inference-time perspective to open-ended creative ideation, where there is no fixed answer distribution to recover, and evaluate two anchorless controls---population-referential instructions and model-proposed semantic directions---against stronger stateful alternatives with the full inference path counted.


\section{Experimental Design}
\subsection{Problem Formulation}
Let $\mathcal{M}$ denote the set of LLMs, $\mathcal{P}$ the set of prompt conditions, $\mathcal{G}$ the set of generation methods, and $\mathcal{S}=\{\texttt{neutral},\texttt{diverge}\}$ the set of instruction strategies. Each prompt condition $p\in\mathcal{P}$ belongs to a task family $f(p)$. A cell is a tuple
$c=(m,p,g,s)\in \mathcal{M}\times\mathcal{P}\times\mathcal{G}\times\mathcal{S}$.

For each cell, the object of evaluation is a candidate pool
$X_c=\{x_{c,1},\ldots,x_{c,n}\}$ with $n=150$. We distinguish between \emph{single-stage} and \emph{two-stage} generation methods. Single-stage methods produce the evaluated pool directly. Two-stage methods generate a seed pool $X_c^{(1)}$ and then produce an evaluated pool $X_c^{(2)}$ conditioned on selected information from that seed pool. The analysis compares evaluated pools, while token accounting counts the full inference path needed to produce them.

The evaluated design contains three LLMs, 12 prompt conditions, six generation methods, two instruction strategies, and 150 outputs per cell. We evaluate each pool along three dimensions: semantic diversity, automatic quality proxies, and full-pipeline inference cost; detailed later.

\subsection{Tasks and Models}
We evaluate three ideation task families: stories ($4$ prompt conditions), alternative uses ($5$), and slogans ($3$). The families differ in output length, genre, and relevant dimensions of variation~\citep{deshpande2025diverse}; see Appendix~\ref{app:tasks-prompts} for full details.

For \textbf{stories}, we use four compact creative-writing prompts. Three are adapted from the \texttt{WritingPrompts} corpus \citep{fan2018hierarchical}: a parachute-failure story, a short horror story, and a life-history/final-seconds microfiction prompt. The fourth is a jungle-adventure prompt adapted from \citet{doshi2024generative}. All story prompts require exactly eight sentences.

For \textbf{alternative uses}, we use the Alternative Uses Task (AUT)~\citep{guildford1978alternate}, a standard paradigm in human and LLM creativity research~\citep{deshpande2025diverse}. Each prompt asks for one alternative use for an everyday object, excluding its common use. The objects are shoe, button, key, wooden pencil, and automobile tire.

For \textbf{slogans}, we use three slogan prompts: a smartphone, a soda, and a blood-donation campaign~\citep{azad2026ex}. Each prompt requires a single slogan of at most six words.

We evaluate GPT-5.4, Claude Sonnet 4.6, and Gemini 2.5 Pro. Because instruction following and contextual use may vary by provider, we report model-specific estimates and use cross-provider comparisons to assess whether design recommendations are qualitatively stable.

\subsection{Instruction Strategies and Generation Methods}
All generation methods are crossed with two instruction strategies. The \texttt{neutral} strategy appends: \emph{``Make the response novel and appropriate for the task.''} The \texttt{diverge} strategy appends the same goal plus: \emph{``Try to make it stand out from other responses that might be generated for this same task.''} For two-stage methods, \texttt{diverge} is contextualized to the seed outputs: the model is asked to produce a response that differs from the previous response(s) shown while still satisfying the original task. We compare two single-stage methods and four two-stage methods:

(i)~\textbf{\texttt{indep}} is single-stage independent generation. The model generates the evaluated pool from the prompt condition alone, without sample ideas, representative anchors, or semantic strata.

(ii)~\textbf{\texttt{strat}} is single-stage semantic direction stratification. It is inspired by the autostratification logic of \textsc{SimpleStrat}~\citep{wong2026simplestrat}, but adapts it for open-ended ideation. \textsc{SimpleStrat} estimates stratum densities and probabilistically samples prompts to better match a target answer distribution; we instead allocate the 150 generation slots \textit{uniformly} across five model-proposed directions, because our goal is to broaden a candidate pool rather than recover the natural distribution of valid answers. We also replace binary, task-specific partition questions with broad semantic direction-based instructions that apply across task families. The pool is generated under direction-specific instructions without any seed examples.

(iii)~\textbf{\texttt{repr}} is two-stage representative avoidance. Following the center-selection strategy in \textsc{G2}~\citep{ruan2025g2}, we embed the seed pool $X_c^{(1)}$ and select three representative anchors using a medoid-start farthest-first rule: the first anchor is the pool medoid, and subsequent anchors maximize distance from already selected anchors. Every second-stage call observes the same three anchors and is asked to differ from them.

(iv)~\textbf{\texttt{self}} is two-stage self anti-anchoring. Each second-stage call observes its own seed response $x_{c,i}^{(1)}$ and regenerates an idea, testing whether the model can use its own prior output as a negative reference~\citep{zhang2024improving}.

(v)~\textbf{\texttt{peer1}} is two-stage dyadic peer anti-anchoring. We partition $X_c^{(1)}$ into 75 dyads. Each second-stage call observes its own seed response and the other response in the dyad. Unlike \texttt{repr}, \texttt{peer1} uses distributed local anti-anchors. 

(vi)~\textbf{\texttt{peer2}} is two-stage triadic peer anti-anchoring. We partition $X_c^{(1)}$ into 50 triads. Each second-stage call observes its own seed response and the other two responses in the triad. Relative to \texttt{peer1}, this increases the amount of local peer contrast available during regeneration.

This design separates two routes to candidate-pool diversification. Single-stage methods test whether diversity can be induced without observed sample ideas, through either a population-referential instruction or model-proposed semantic directions. Two-stage methods test whether observed seed outputs can redirect regeneration through shared representative anti-anchors or distributed local anti-anchors.

\subsection{Diversity Metrics}\label{diversity_metrics}
We represent each output $x$ with a normalized sentence embedding $f(x)$ and define semantic distance as $d_{ij}=1-\cos(f(x_i),f(x_j))$. The primary metric is mean pairwise semantic distance:
\[
D_{\texttt{pair}}(X_c)=\frac{2}{n(n-1)}\sum_{i<j} d_{ij}.
\]

We use four additional diversity summaries for robustness. Mean nearest-neighbor distance,
\[
D_{\texttt{nn}}(X_c)=\frac{1}{n}\sum_i \min_{j\neq i} d_{ij},
\]
captures local redundancy. Mean medoid distance,
\[
D_{\texttt{med}}(X_c)=\frac{1}{n}\sum_i d_{ir},
\qquad
r=\arg\min_\ell \sum_j d_{\ell j},
\]
captures dispersion around the pool's most central output. MST dispersion,
\[
D_{\texttt{mst}}(X_c)=\frac{1}{n-1}\sum_{(i,j)\in T_c} d_{ij},
\]
is the average edge length of the minimum spanning tree $T_c$ under $d_{ij}$, capturing sparse global spread. Finally, normalized region entropy measures how outputs are distributed across prompt-level semantic regions:
\[
D_{\texttt{ent}}(X_c)=
-\frac{1}{\log K_p} \sum_k \pi_{c,k}\log \pi_{c,k},
\]
where $\pi_{c,k}$ is the share of outputs from cell $c$ assigned to region $k$. Here, outputs within each prompt condition are clustered with K-means in embedding space. The denominator $\log K_p$ uses the fixed prompt-level region count, so the metric rewards both occupying more regions and distributing outputs more evenly across them. See Appendix~\ref{supplementary_notes} for further notes.

\subsection{Quality Proxies}
We use task-specific automatic proxies to test whether diversification trades off against output quality. Stories are scored with MAoSS~\citep{luchini2025automated}; AUT responses are scored with CLAUS~\citep{patterson2023multilingual}, both validated automatic measures for their respective task formats. Slogans are scored with a phrase-level non-template metric~\citep{azad2026ex}, which measures leave-one-out bigram and trigram reuse within each slogan task, with higher values indicating less reliance on repeated phrase templates. This metric is appropriate for slogans because a central quality risk in short marketing phrases is boilerplate recombination rather than length or elaboration.

Because slogan scoring is less standardized than AUT or story scoring, we use two robustness checks: replacing the lexical slogan score with an LLM-as-a-judge creativity score, and omitting slogans entirely while recomputing all quality-bearing analyses over AUT and story tasks only. Because the raw proxy scales are not comparable across task families, we standardize quality within task: $Q^z(x)=\frac{Q(x)-\mu_t}{\sigma_t}$,
where $\mu_t$ and $\sigma_t$ are computed over all evaluated outputs for task $t$. Cell-level quality is the mean standardized score across outputs.

\subsection{Pool-Size Rarefaction}
To examine how diversity accumulates as a pool is sampled, we perform pool-size rarefaction. For each $X_c$, we repeatedly sample subpools $X_{c,q}\subset X_c$ without replacement for $q=1,\ldots,150$ and compute $D_{\texttt{pair}}(X_{c,q})$ and $D_{\texttt{ent}}(X_{c,q})$.

For each metric, we summarize the rarefaction curve with AUC and first-hit. AUC measures average diversity across subpool sizes. First-hit is the smallest $q$ needed to match the corresponding metric value of the full 150-output \texttt{indep}--\texttt{neutral} pool for the same model and prompt condition. Rarefaction is computed within already generated pools; it is not adaptive generation.

\subsection{Token Accounting}
We report full-pipeline token cost for each evaluated pool. \texttt{indep} counts only single-stage generation. \texttt{strat} counts the planning call plus single-stage stratum-guided generation. The two-stage methods---\texttt{repr}, \texttt{self}, \texttt{peer1}, and \texttt{peer2}---count both $X_c^{(1)}$ and $X_c^{(2)}$. For each cell, we report the token multiplier relative to the single-stage neutral baseline for the same model and prompt condition:
\[
R^{\mathrm{tok}}_{m,p,g,s}
=
\frac{\mathrm{Tok}_{m,p,g,s}}
{\mathrm{Tok}_{m,p,\texttt{indep},\texttt{neutral}}}.
\]
This makes the comparison sensitive to overhead introduced by planning, seed-pool generation, and second-stage regeneration.

\subsection{Contrasts, Aggregation, and Uncertainty}
Let $Y_{m,p,g,s}$ denote any cell-level outcome. We report two planned contrasts. The baseline contrast compares each configuration to \texttt{indep}--\texttt{neutral} within the same model and prompt condition:
\[
\Delta^{\mathrm{base}}_{m,p,g,s}(Y)
=
Y_{m,p,g,s}
-
Y_{m,p,\texttt{indep},\texttt{neutral}}.
\]
The divergence contrast isolates the marginal effect of \texttt{diverge} within each generation method:
\[
\Delta^{\mathrm{div}}_{m,p,g}(Y)
=
Y_{m,p,g,\texttt{diverge}}
-
Y_{m,p,g,\texttt{neutral}}.
\]

All quantities are computed within prompt condition and aggregated using task-family design averages: we average equally across prompt conditions within stories, AUT, and slogans, then report model-specific family averages. We report percentile intervals from output-level bootstrap resampling for full-pool metrics and repeated subpool sampling for rarefaction curves. Because the central question is whether design recommendations hold across models, we emphasize model-specific estimates rather than a single pooled provider effect.


\begin{table}[t]
\centering
\small
\setlength{\tabcolsep}{4.5pt}
\begin{tabular}{lccc}
\toprule
Model &
$\Delta^{\mathrm{div}}D_{\texttt{pair}}$ &
$\Delta^{\mathrm{div}}Q^z$ &
$R^{\mathrm{tok}}$ \\
\midrule
Cl. Son. 4.6 &
0.06 [0.05, 0.07] &
0.38 [0.34, 0.43] &
1.1 \\
Gem 2.5 Pro &
0.05 [0.04, 0.06] &
0.31 [0.24, 0.37] &
1.1 \\
GPT-5.4 &
0.04 [0.03, 0.05] &
0.25 [0.20, 0.30] &
1.1 \\
\bottomrule
\end{tabular}
\caption{Effect of \texttt{diverge} under \texttt{indep}. Values are task-family design averages with 95\% bootstrap CI.}
\label{tab:indep-diverge-lift}
\end{table}
\begin{figure*}[t]
\centering
\includegraphics[width=\linewidth]{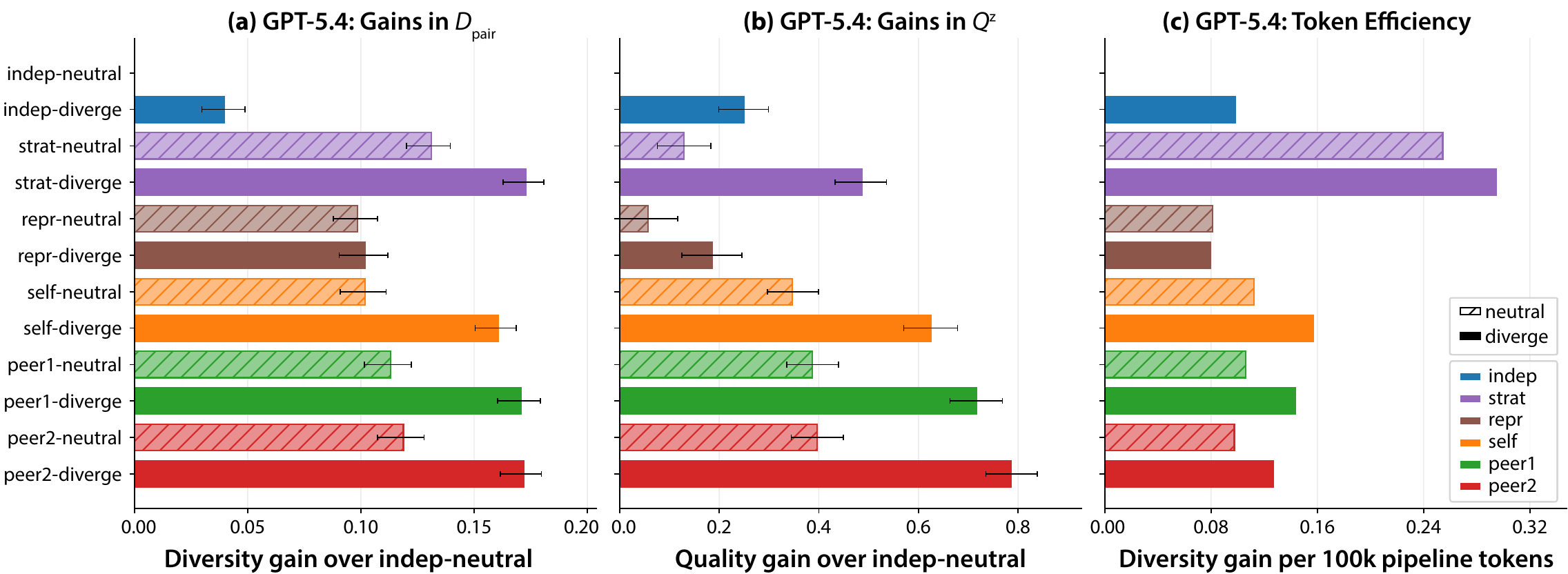}
\caption{GPT-5.4 diversity--quality--efficiency summary. Panels (a) and (b) report $\Delta^{\mathrm{base}}$ contrasts relative to \texttt{indep}--\texttt{neutral} within the same model and prompt condition, with 95\% bootstrap intervals. Panel (c) reports token-normalized diversity gain, computed as $D_{\texttt{pair}}$ gain per 100k full-pipeline tokens. Colors match generation methods, with neutral and \texttt{diverge} variants shown in matched pairs. Provider-specific analogues and all-provider token-efficiency summaries are reported in Appendix~\ref{app:supplementary-figures}.}
\label{fig:gpt54-frontier}
\end{figure*}

\section{Results}
\subsection{A Direct Divergence Instruction Is a Strong Low-Cost Baseline}

We examine \texttt{indep} under \texttt{diverge}, the lowest-cost diversification control in our design. The model generates from the task prompt alone, and the only change from \texttt{indep}--\texttt{neutral} is the population-referential instruction to stand out from other responses that might be generated for the same task.

For all three models, \texttt{diverge} improves both $D_{\texttt{pair}}$ and $Q^z$ under independent generation (Table~\ref{tab:indep-diverge-lift}). The diversity lift ranges from 0.0392 [0.0299, 0.0486] for GPT-5.4 to 0.0622 [0.0534, 0.0722] for Claude Sonnet 4.6; quality-proxy gains are also positive for all providers. Because \texttt{indep}--\texttt{diverge} remains single-stage, its token cost is close to the neutral baseline: $R^{\mathrm{tok}}\approx1.1$ for all three models. In Figure~\ref{fig:gpt54-frontier}, it is not the highest-diversity method, but it occupies the low-cost region of the design space.

This makes \texttt{indep}--\texttt{diverge} the low-cost baseline that more elaborate inference-time diversification methods should exceed.

\subsection{Semantic Direction Stratification Provides a Strong Anchorless Diversification Signal}
\label{sec:results-stratification}
We next examine the full set of generation methods. The methods differ in how much information they use at inference time. \texttt{indep} and \texttt{strat} are anchorless: they do not observe generated candidate ideas. In contrast, \texttt{self}, \texttt{peer1}, \texttt{peer2}, and \texttt{repr} are two-stage methods that first generate seed outputs and then use those outputs as anti-anchors. This comparison asks whether example-free controls can approach the diversity gains of methods that observe and react to prior generations.

Most non-baseline configurations improve $D_{\texttt{pair}}$ relative to \texttt{indep}--\texttt{neutral}. For GPT-5.4, the largest gains are tightly grouped: \texttt{strat}--\texttt{diverge} improves $D_{\texttt{pair}}$ by 0.1717 [0.1624, 0.1810], \texttt{peer2}--\texttt{diverge} by 0.1706 [0.1612, 0.1794], and \texttt{peer1}--\texttt{diverge} by 0.1695 [0.1598, 0.1780]. Thus, for GPT-5.4, \texttt{strat}--\texttt{diverge} reaches the same top tier as the strongest peer-anchored methods but is not clearly separated from them.

The comparison is more favorable to \texttt{strat} for the other providers. For Claude Sonnet 4.6, \texttt{strat}--\texttt{diverge} gives the largest observed diversity gain, 0.2526 [0.2443, 0.2610], compared with 0.2081 [0.1999, 0.2167] for the strongest peer-anchored condition. For Gemini 2.5 Pro, the corresponding gains are 0.1335 [0.1231, 0.1443] and 0.1278 [0.1172, 0.1378]. These results indicate that model-proposed semantic directions can supply a strong diversification signal even without examples.

Quality-proxy patterns are more heterogeneous. For GPT-5.4 and Gemini 2.5 Pro, the largest $Q^z$ gains occur under \texttt{peer2}--\texttt{diverge}; for Claude Sonnet 4.6, the largest gain occurs under \texttt{strat}--\texttt{diverge}. Thus, no method is uniformly dominant; \texttt{strat}--\texttt{diverge} is consistently competitive on quality while being among the strongest configurations for diversity and remaining single-stage and anchorless.

The divergence instruction also composes with most generation methods. $\Delta^{\mathrm{div}}_{m,p,g}$ is positive for \texttt{indep}, \texttt{self}, \texttt{peer1}, \texttt{peer2}, and \texttt{strat} for all three models; \texttt{repr} is mostly inconclusive. Within the anchorless subset, \texttt{strat}--\texttt{diverge} is the strongest configuration in this design. The divergence lift under \texttt{strat} is 0.04 [0.03, 0.05] for GPT-5.4, 0.04 [0.03, 0.05] for Claude Sonnet 4.6, and 0.025 [0.02, 0.03] for Gemini 2.5 Pro.

\subsection{Pool-Size Rarefaction Analysis}
\label{sec:results-rarefaction}
\begin{figure}[t]
\centering
\includegraphics[width=\linewidth]{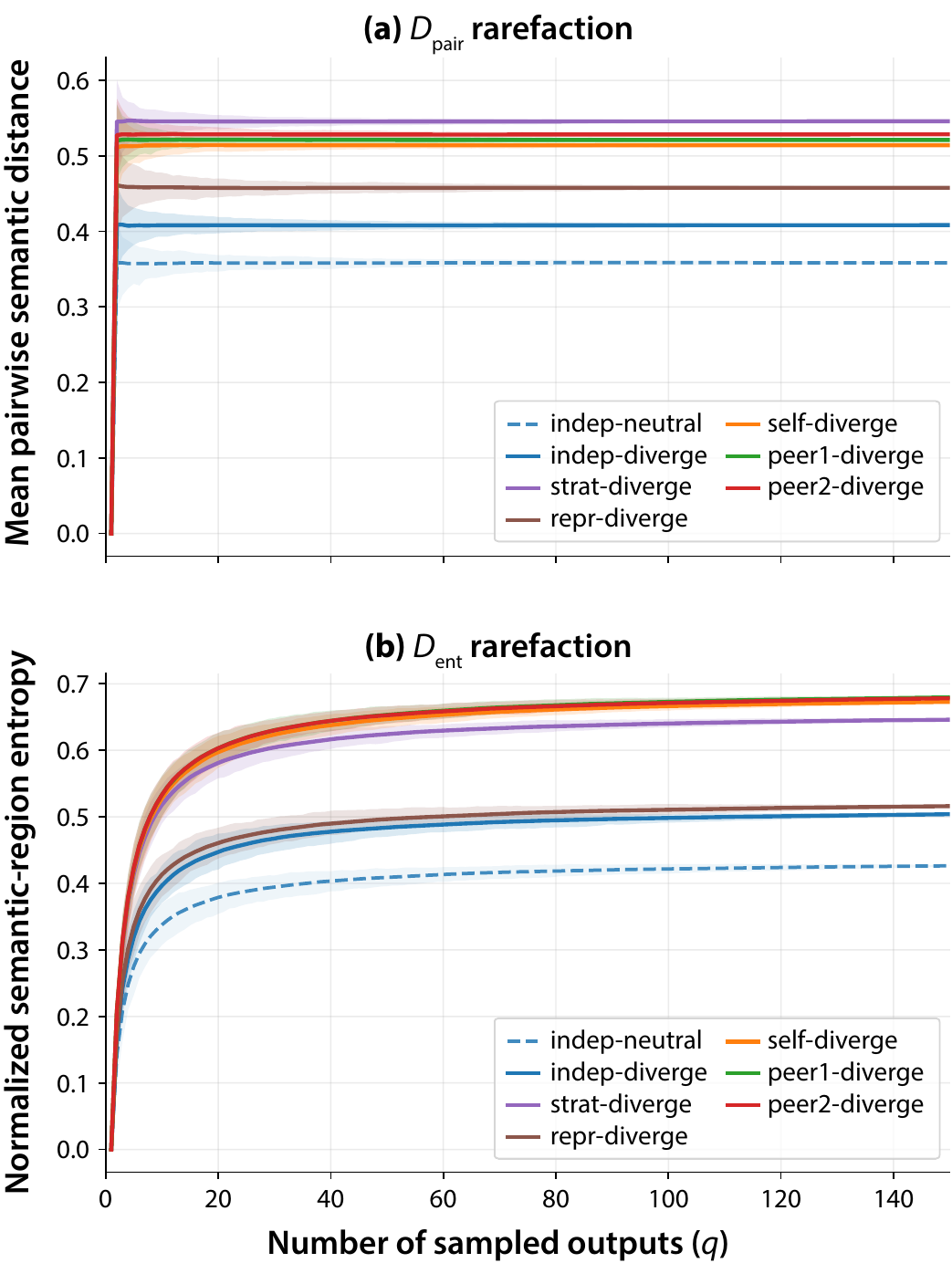}
\caption{Rarefaction curves for the \texttt{indep}--\texttt{neutral} baseline and all six \texttt{diverge} configurations. Panel (a) shows average pairwise semantic distance, $D_{\texttt{pair}}$, and panel (b) shows semantic-region entropy, $D_{\texttt{ent}}$. Shaded bands indicate 95\% bootstrap intervals.}
\label{fig:rarefaction-main}
\end{figure}

The diversity gains in Section~\ref{sec:results-stratification} raise two related questions. First, do higher values of $D_{\texttt{pair}}$ reflect broader semantic spread, or could a method move outputs away from the neutral baseline while concentrating them elsewhere? Second, when a method does broaden the pool, how quickly does that breadth accumulate as outputs are sampled? We address both questions with pool-size rarefaction over the already generated pools. Figure~\ref{fig:rarefaction-main} shows the resulting trajectories for $D_{\texttt{pair}}$ and semantic-region entropy.

Pairwise rarefaction mirrors the full-pool analysis. Relative to \texttt{indep}--\texttt{neutral}, \texttt{strat}--\texttt{diverge} increases $D_{\texttt{pair}}$ AUC by 0.187 [0.091, 0.328] and exceeds \texttt{repr}--\texttt{diverge} by 0.088 [0.004, 0.294]. Because $D_{\texttt{pair}}$ stabilizes quickly as $q$ grows, the main information in Figure~\ref{fig:rarefaction-main}(a) is the persistent vertical separation between methods.

Entropy rarefaction gives a clearer view of distributional spread (Figure~\ref{fig:rarefaction-main}(b)). Local anchor methods accumulate semantic-region entropy slightly faster than \texttt{strat}--\texttt{diverge}: the largest design-average entropy AUCs are obtained by \texttt{peer1}--\texttt{diverge} and \texttt{peer2}--\texttt{diverge} (both 0.64 after rounding), followed by \texttt{self}--\texttt{diverge} (0.63) and \texttt{strat}--\texttt{diverge} (0.61). This is intuitive, as local anchors give each second-stage call concrete prior outputs to move away from.

The anchorless comparison remains favorable to \texttt{strat}. Relative to \texttt{indep}--\texttt{neutral}, \texttt{strat}--\texttt{diverge} increases entropy AUC by 0.210 [0.079, 0.346]; relative to \texttt{repr}--\texttt{diverge}, it increases entropy AUC by 0.125 [0.014, 0.267]. It reaches the full \texttt{indep}--\texttt{neutral} entropy target after 7.4 sampled outputs on average, compared with 17.9 for \texttt{indep}--\texttt{diverge} and 23.6 for \texttt{repr}--\texttt{diverge}. Local anchor methods reach the target slightly earlier, with means between 6.6 and 7.2 outputs, but require a seed pool for producing the evaluated pool.

Together, rarefaction clarifies the tradeoff: local self- and peer-anchor methods achieve the fastest entropy accumulation among evaluated pools, but they require seed outputs. Semantic direction stratification nearly matches this accumulation while remaining anchorless, and substantially outperforms independent divergence and shared representative avoidance on both $D_{\texttt{pair}}$ and $D_{\texttt{ent}}$ AUC. We next ask whether these diversity gains remain attractive under full-pipeline token accounting.

\subsection{Full-Pipeline Token Accounting}
\label{sec:results-token-accounting}
Full-pipeline token accounting compares the inference paths needed to produce evaluated pools. \texttt{indep} counts only single-stage generation; \texttt{strat} adds a planning call; the two-stage methods require both seed-pool and evaluated-pool generation.

For GPT-5.4, \texttt{peer2}--\texttt{diverge} and \texttt{strat}--\texttt{diverge} produce nearly identical $D_{\texttt{pair}}$ gains (0.1719 versus 0.1730), but their costs differ: $R^{\mathrm{tok}}=3.71$ for \texttt{peer2}--\texttt{diverge} and $1.61$ for \texttt{strat}--\texttt{diverge}. Consequently, \texttt{strat}--\texttt{diverge} yields the largest GPT-5.4 diversity gain per 100k pipeline tokens (0.295), followed by \texttt{strat}--\texttt{neutral} (0.254) and \texttt{self}--\texttt{diverge} (0.157). The same top ranking holds for Claude Sonnet 4.6 (0.379) and Gemini 2.5 Pro (0.207).

Quality-normalized accounting is similar. In absolute $Q^z$, local peer anchoring is strongest: \texttt{peer2}--\texttt{diverge} gives the largest raw quality-proxy gain for GPT-5.4 and Gemini 2.5 Pro. After normalizing by full-pipeline tokens, however, \texttt{strat}--\texttt{diverge} gives the largest $Q^z$ gain per 100k tokens for all three providers: 0.829 for GPT-5.4, 1.211 for Claude Sonnet 4.6, and 0.783 for Gemini 2.5 Pro.

Figure~\ref{fig:gpt54-frontier} illustrates the tradeoff for GPT-5.4; provider analogues appear in Appendix~\ref{app:supplementary-figures}. The practical pattern is stable: \texttt{indep}--\texttt{diverge} is the cheapest nontrivial intervention, while \texttt{strat}--\texttt{diverge} occupies the strongest diversity--quality--efficiency region in this design. Because provider tokenizers and accounting conventions differ, token-normalized comparisons are interpreted within the provider rather than as absolute cross-provider cost comparisons.

\subsection{Robustness}
\label{sec:results-robustness}
Across alternate full-pool diversity metrics, the same structure recurs. Relative to \texttt{indep}--\texttt{neutral}, \texttt{strat}--\texttt{diverge} gives the largest design-average gains in $D_{\texttt{pair}}$ (0.187), $D_{\texttt{med}}$ (0.173), and full-pool $D_{\texttt{ent}}$ (0.220). Local anchor methods lead on spacing-sensitive diagnostics: \texttt{peer2}--\texttt{diverge} gives the largest gains in $D_{\texttt{nn}}$ (0.093) and $D_{\texttt{mst}}$ (0.087), followed closely by \texttt{peer1}--\texttt{diverge} and \texttt{self}--\texttt{diverge}. The representative anti-anchor condition is weaker on region-level spread: \texttt{repr}--\texttt{diverge} improves over the baseline but has a smaller entropy gain (0.090) than \texttt{repr}--\texttt{neutral} (0.120), consistent with the rarefaction result that fixed shared anchors move outputs away from examples without producing comparable semantic-region coverage (see tables in Appendix~\ref{app:tables}).

The quality checks give the same reading. Replacing the lexical slogan score with an LLM-as-a-judge creativity score preserves the main pattern: \texttt{diverge} remains quality-preserving or quality-improving, and \texttt{strat}--\texttt{diverge} remains on the favorable diversity--quality--compute frontier. Dropping slogans entirely yields the same conclusion over AUT and story tasks alone: \texttt{diverge} increases quality within every generation method, with bootstrap intervals excluding zero. The one qualification is narrow: \texttt{strat}--\texttt{neutral} no longer shows a reliable quality gain without slogans, while \texttt{strat}--\texttt{diverge} remains clearly positive. Overall, the quality conclusions remain robust to metric choice.
 

\section{Discussion}
This paper studied anchorless diversification for parallel LLM ideation. Across three creative task families and three frontier models, we find that models can be steered toward broader candidate pools through inference-time controls that preserve the appeal of parallel generation. A population-referential divergence instruction provides the simplest control: it increases semantic diversity while preserving or improving automatic quality proxies at nearly the same token cost as independent generation. Semantic direction stratification is stronger. With a single planning call, the model externalizes broad task-specific directions and then generates candidates within them, producing pools competitive with or stronger than seed-dependent regeneration methods. Under full-pipeline token accounting, stratified divergence yields the most favorable diversity--quality--compute frontier in our design.

These findings sharpen the inference-time diversification literature. Prior work has increased diversity through decoding controls, prompt variation, critique, voting, self-improvement, evolutionary search, and anti-anchoring against prior generations~\citep{vijayakumar2016diverse,holtzman2019curious,chung2023increasing,lahoti2023improving,franceschelli2024creative,bradley2024quality,hayati2024far,ruan2025g2}. We add a deployment-centered comparison: methods are evaluated not only by returned-pool diversity, but by the complete inference process required to obtain it. Candidate-pool generation is useful only when diversity gains survive the costs of planning, seed generation, selection, and regeneration. Anchored regeneration remains powerful, but anchorless stratification offers a better balance when the objective is broad parallel exploration under realistic inference budgets.

The results also support an autostratification view of LLM ideation. Rather than treating the model only as a sampler from a prompt-conditioned distribution, stratification asks it to first articulate a coarse map of the response space. This map is imperfect, but useful: it exposes semantically meaningful regions that default sampling may underexplore and converts the model's latent sense of task variation into an explicit allocation structure. Autostratified generation can remain stateless after planning, parallelize cleanly across directions, and avoid dependence on previously observed ideas.

These findings have broad implications for applications where LLMs generate candidate pools for downstream selection. For instance, in AI-scientist systems, autonomous agents increasingly propose research questions, experiments, and manuscripts. In drug discovery, generative models can propose molecular candidates, mechanisms, or experimental directions that may appear promising individually while crowding around familiar regions of the search space. Without diversification, such systems can amplify AI-induced idea diversity collapse: many users or agents receive plausible but crowded candidates from the same semantic or conceptual basins. Anchorless controls may help preserve the portfolio value of parallel generation itself.


\section{Limitations}

This study has several limitations. First, our quality analyses rely on automatic proxies. MAoSS and CLAUS are validated for stories and alternative uses, and we include robustness checks for slogans using an LLM-as-a-judge score, but these measures do not replace expert or user evaluation. In particular, semantic diversity can be valuable only when the resulting candidates remain useful for downstream selection, refinement, or recombination. Future work should test whether the diversified pools identified here improve human choice, expert-rated creativity, or downstream agentic development.

Second, our diversity measures are embedding-based. Pairwise distance, nearest-neighbor distance, medoid distance, MST dispersion, and semantic-region entropy capture complementary aspects of pool structure, but they remain sensitive to the embedding model and clustering choices. These metrics are appropriate for comparing methods under a common representation, yet they may miss task-specific distinctions that matter to human judges.

Third, the experiments cover three creative task families and three contemporary LLMs. The consistency of the main pattern across providers is encouraging, but the results should not be interpreted as a universal ranking of methods. Other domains, longer-form outputs, stricter factual constraints, multimodal tasks, or future model families may change the balance between anchorless stratification and seed-dependent regeneration.

Finally, our design fixes several implementation choices: five semantic directions, equal allocation across directions, 150 outputs per pool, and a single form of population-referential divergence instruction. These choices make the comparison clean, but they leave open whether adaptive allocation, alternative direction counts, richer planning prompts, or hybrid stratification-and-regeneration pipelines could further improve the diversity--quality--compute frontier.

\section{Ethical Considerations}
We analyzed LLM-generated data under carefully curated prompts and did not collect any new human data for this research. Given the nature of the research in creative diversity, we do not readily foresee potential harm or risk of our contributions.

\section*{Acknowledgments}

A faculty startup fund at the University of South Florida supported this work.

\section*{Code Availability}
The full codebase for collecting and processing the LLM API data is available here: \url{https://github.com/cssai-research/anchorless-diversification}.


\bibliography{references}


\appendix

\section{Task and Prompt Details}
\label{app:tasks-prompts}
This appendix reports the prompt structure used for model generation. Each model call produced exactly one output. Unless otherwise noted, calls used the following role structure: \texttt{SYSTEM:} shared system instruction and \texttt{USER:} task-specific instructions + instruction-strategy modifier + generation-method instructions.

The task-specific instructions instantiate the prompt condition $p$, the instruction-strategy modifier instantiates $s\in\{\texttt{neutral},\texttt{diverge}\}$, and the generation-method instructions instantiate $g$. For \texttt{indep}, the generation-method component is empty. For \texttt{strat}, it gives the assigned semantic stratum. For \texttt{repr}, \texttt{self}, \texttt{peer1}, and \texttt{peer2}, it gives the prior-output context used for second-stage generation. Duplicate generations were retained. No semantic filtering or manual editing was applied to generated content.

\subsection{Shared \texttt{SYSTEM} Instruction}
All ordinary generation calls used the same \texttt{SYSTEM} instruction:

\begin{quote}
\small
You are participating in a controlled creativity experiment. Follow the task instructions exactly. Return exactly one response. Do not explain your reasoning. Do not include commentary before or after the response.
\end{quote}

The only exception is the \texttt{strat} planning call, which used the planning-specific \texttt{SYSTEM} instruction reported in Section~\ref{app:strat-prompt}.

\subsection{Task-Specific Instructions}

In ordinary generation calls, task-specific instructions came as the first segment of \texttt{USER}.

\subsubsection{Story Prompts}

All story prompts required exactly eight sentences, English text, and content appropriate for a teenage or young-adult audience. This standardization reduced uncontrolled variation in output length.

\paragraph{Jungle adventure.}
\begin{quote}
\small
You are participating in a creative writing task.

Write exactly one story about an adventure in the jungle.

Requirements:

- The story must be exactly 8 sentences long.

- The story must be written in English.

- The story must be appropriate for a teenage and young adult audience, approximately ages 15 to 24.

- Do not provide multiple story ideas.

- Do not summarize the story.

- Return only the story.
\end{quote}

\paragraph{Parachute failure.}
\begin{quote}
\small
You are participating in a creative writing task.

Write exactly one story based on this prompt:

The parachute isn’t opening up.

Requirements:

- The story must be exactly 8 sentences long.

- The story must be written in English.

- The story must be appropriate for a teenage and young adult audience, approximately ages 15 to 24.

- Do not provide multiple story ideas.

- Do not summarize the story.

- Return only the story.
\end{quote}

\paragraph{Short horror.}
\begin{quote}
\small
You are participating in a creative writing task.

Write exactly one short horror story designed to chill the bones.

Requirements:

- The story must be exactly 8 sentences long.

- The story must be written in English.

- The story must be appropriate for a teenage and young adult audience, approximately ages 15 to 24.

- Do not provide multiple story ideas.

- Do not summarize the story.

- Return only the story.
\end{quote}

\paragraph{Life and final seconds.}
\begin{quote}
\small
You are participating in a creative writing task.

Write exactly one story in 8 sentences. The first sentence must describe 100 years of a character's life. The next 7 sentences must describe the last 10 seconds of that character's life.

Requirements:

- The story must be exactly 8 sentences long.

- The story must be written in English.

- The story must be appropriate for a teenage and young adult audience, approximately ages 15 to 24.

- Do not number or label the sentences.

- Do not state which sentence does what.

- Return only the story as one paragraph.
\end{quote}

\subsubsection{Alternative Uses Prompts}
All Alternative Uses Task prompts used the same template, varying only the object and the common use to avoid (Table~\ref{tab:appendix-aut}).

\begin{quote}
\small
You are participating in a creativity task.

Object: \textless object\textgreater

Common use to avoid: \textless common use\textgreater

Generate exactly one unusual, novel, and plausible alternative use for the object or one of its parts.

Requirements:

- Do not use the common use.

- Do not list multiple uses.

- The response must be written in English.

- Return only the alternative use as a short phrase or one sentence.
\end{quote}

\begin{table}[h]
\centering
\small
\setlength{\tabcolsep}{4pt}
\begin{tabular}{ll}
\toprule
Object & Common use to avoid \\
\midrule
shoe & used as footwear \\
button & used to fasten things \\
key & used to open a lock \\
wooden pencil & used for writing \\
automobile tire & used on the wheel of an automobile \\
\bottomrule
\end{tabular}
\caption{Alternative Uses Task object conditions.}
\label{tab:appendix-aut}
\end{table}

\subsubsection{Slogan Prompts}
All slogan prompts required exactly one English slogan of at most six words.

\paragraph{Smartphone slogan.}
\begin{quote}
\small
You are part of the marketing team at a tech company preparing to launch a new smartphone.

Generate exactly one marketing slogan for this brand-new smartphone.

Requirements:

- The slogan must not exceed 6 words.

- The slogan must be written in English.

- You may assume any detail about the smartphone.

- Do not list multiple slogans.

- Return only the slogan text.
\end{quote}

\paragraph{Soda slogan.}
\begin{quote}
\small
You are part of the marketing team at a beverage company preparing to launch a new soda.

Generate exactly one marketing slogan for this brand-new soda.

Requirements:

- The slogan must not exceed 6 words.

- The slogan must be written in English.

- You may assume any detail about the soda.

- Do not list multiple slogans.

- Return only the slogan text.
\end{quote}

\paragraph{Blood donation campaign slogan.}
\begin{quote}
\small
You are part of the communications team at a nonprofit organization preparing a campaign to encourage blood donation.

Generate exactly one campaign slogan for this blood donation campaign.

Requirements:

- The slogan must not exceed 6 words.

- The slogan must be written in English.

- You may assume any detail about the campaign.

- Do not list multiple slogans.

- Return only the slogan text.
\end{quote}

\subsection{Instruction-Strategy Modifiers}

Each ordinary generation \texttt{USER} prompt included exactly one of the following instruction-strategy modifiers, placed after the task-specific instructions.

\paragraph{\texttt{neutral}.}
\begin{quote}
\small
Creativity goal:

- Make the response novel and appropriate for the task.
\end{quote}

\paragraph{\texttt{diverge}.}
\begin{quote}
\small
Creativity goal:

- Make the response novel and appropriate for the task.

- Try to make it stand out from other responses that might be generated for this same task.
\end{quote}

For second-stage methods that showed prior outputs, the same modifier was included, and the final method-specific sentence was also adapted to the displayed context, as reported in Section~\ref{app:method-instructions}.

\subsection{Generation-Method Instructions}
\label{app:method-instructions}

This subsection reports the generation-method component appended after the task-specific instructions and instruction-strategy modifier. 

\subsubsection{\texttt{indep}: Independent Direct Generation}

For \texttt{indep}, the generation-method component was empty. The direct pool $X_c^{(1)}$ was generated from the task-specific instructions and instruction-strategy modifier alone.

\subsubsection{\texttt{repr}, \texttt{self}, \texttt{peer1}, and \texttt{peer2}: First-Stage Generation}

For \texttt{repr}, \texttt{self}, \texttt{peer1}, and \texttt{peer2}, the first-stage seed pool $X_c^{(1)}$ was generated using the same prompt form as \texttt{indep}. The generation-method component was empty in the first stage. Anti-anchoring affected only the second-stage context (described below).

\subsubsection{\texttt{self}: Second-Stage Self Anti-Anchoring}
For each seed response $x_{c,i}^{(1)}$, the following context was appended:

\begin{quote}
\small
Previous response from your first round:

``\textless self response\textgreater''

Now generate one new response for the same task.
\end{quote}

Under \texttt{diverge}, the final sentence was replaced with:

\begin{quote}
\small
Now generate one new response for the same task. It should stand out from the previous response(s) shown above while still satisfying all task requirements.
\end{quote}

The resulting outputs form $X_c^{(2)}$.

\subsubsection{\texttt{peer1}: Second-Stage Dyadic Peer Anti-Anchoring}

For \texttt{peer1}, the seed pool was partitioned into 75 dyads. For each seed response $x_{c,i}^{(1)}$, the following context was appended:

\begin{quote}
\small
Previous response from your first round:

``\textless self response\textgreater''

Previous response from another agent in the same first round:

``\textless peer response\textgreater''

Now generate one new response for the same task.
\end{quote}

Under \texttt{diverge}, the final sentence was replaced with:

\begin{quote}
\small
Now generate one new response for the same task. It should stand out from the previous response(s) shown above while still satisfying all task requirements.
\end{quote}

The resulting outputs form $X_c^{(2)}$.

\subsubsection{\texttt{peer2}: Second-Stage Triadic Peer Anti-Anchoring}

For \texttt{peer2}, the seed pool was partitioned into 50 triads. For each seed response $x_{c,i}^{(1)}$, the following context was appended:

\begin{quote}
\small
Previous response from your first round:

``\textless self response\textgreater''

Previous responses from two other agents in the same first round:

1. ``\textless peer response 1\textgreater''

2. ``\textless peer response 2\textgreater''

Now generate one new response for the same task.
\end{quote}

Under \texttt{diverge}, the final sentence was replaced with:

\begin{quote}
\small
Now generate one new response for the same task. It should stand out from the previous response(s) shown above while still satisfying all task requirements.
\end{quote}

The resulting outputs form $X_c^{(2)}$.

\subsubsection{\texttt{repr}: Second-Stage Representative Avoidance}

The first-stage seed pool was embedded using \texttt{sentence-transformers/all-mpnet-base-v2}, and three anchors were selected using a medoid-start farthest-first rule: the first anchor was the pool medoid, and subsequent anchors were selected to maximize distance from the anchors already selected.

For second-stage \texttt{repr} generation, the ordinary shared \texttt{SYSTEM} instruction, task-specific instructions, and instruction-strategy modifier were used, and the following generation-method component was appended. The same three representative anchors were shown to all 150 final calls for the corresponding model and prompt condition.

\begin{quote}
\small
Previous responses from three other agents in the same first round:

1. ``\textless representative response 1\textgreater''

2. ``\textless representative response 2\textgreater''

3. ``\textless representative response 3\textgreater''

Now generate one new response for the same task.
\end{quote}

Under \texttt{diverge}, the final sentence was replaced with:

\begin{quote}
\small
Now generate one new response for the same task. It should stand out from the previous response(s) shown above while still satisfying all task requirements.
\end{quote}

The resulting outputs form $X_c^{(2)}$.

\subsubsection{\texttt{strat}: Planning-Guided Direct Generation}
\label{app:strat-prompt}

The \texttt{strat} method used a planning call followed by stratum-guided direct generation. The planning call did not use the shared \texttt{SYSTEM} instruction and did not include a \texttt{neutral}/\texttt{diverge} modifier. Its role structure was: \texttt{SYSTEM:} planning system instruction, and \texttt{USER:} planning instructions containing the task-specific instructions.

The planning \texttt{SYSTEM} instruction was:

\begin{quote}
\small
You identify semantic diversity strata for controlled text-generation experiments. Return valid JSON only. Do not include markdown fences or commentary.
\end{quote}

The planning \texttt{USER} prompt was:

\begin{quote}
\small
We will later generate 150 independent responses to the following task.

Task prompt:

\textless task-specific instructions\textgreater

Identify exactly 5 mutually distinct semantic strata for valid responses to this task.

Use the following procedure internally before choosing the final strata:

1. Consider questions that would separate the space of possible valid responses into broad, meaningfully different groups.

2. Prefer distinctions that would split the possible valid responses into reasonably balanced groups, rather than isolating rare edge cases.

3. Convert the best distinctions into categorical conceptual directions for generation.

4. Exclude distinctions based only on superficial wording, tone, length, punctuation, formatting, or synonyms.

5. Exclude strata that name a specific candidate answer, force a specific phrase, or make the original task harder to satisfy.

The final strata must satisfy all of these requirements:

- Each stratum must be a semantic/content direction, not a superficial style change.

- Each stratum must be broad enough to support many different valid responses.

- The strata must be mutually distinct enough that responses generated under different strata are likely to differ conceptually.

- The strata must collectively cover a wide range of plausible valid responses to the task.

- The generation\_instruction must be concise and usable as an added constraint in a later generation prompt.

Return JSON only with this exact structure:

\{
  "task\_id": "...",
  "strata": [
    \{
      "stratum\_id": 1,
      "name": "short name",
      "description": "one sentence describing the semantic direction",
      "generation\_instruction": "a concise instruction that can be appended to the generation prompt",
      "why\_broad": "one short sentence explaining why this stratum can support many valid responses",
      "why\_distinct": "one short sentence explaining how this stratum differs from the other strata"
    \}
  ]
\}

The JSON must contain exactly 5 strata with stratum\_id values 1 through 5.

Do not include any text outside the JSON.
\end{quote}

The returned JSON was parsed and validated to contain exactly five strata with the required fields. Final generation slots were assigned cyclically across strata, yielding 30 generations per stratum. For each final \texttt{strat} call, the ordinary shared \texttt{SYSTEM} instruction was restored, the task-specific instructions and instruction-strategy modifier were included, and the following generation-method component was appended:

\begin{quote}
\small
Conceptual direction assigned for this round:

\textless stratum name\textgreater: \textless stratum description\textgreater

Additional generation constraint:

\textless stratum generation instruction\textgreater

Use this direction as the main conceptual path for the response. Do not mention the direction label or explain the direction.

Now generate one new response for the same task.
\end{quote}

Under \texttt{diverge}, the final sentence was replaced with:

\begin{quote}
\small
Now generate one new response for the same task. It should stand out from other responses that might be generated for this same task while still satisfying all task requirements.
\end{quote}

No seed-output examples were shown in \texttt{strat}; the resulting outputs form a direct pool $X_c^{(1)}$.

\subsection{Generation Parameters}
For ordinary generation calls, each request asked for one candidate response. The generation temperature was $1.0$. Maximum output-token ceilings were set by task family: 2048 for stories, 768 for alternative uses, and 512 for slogans. The \texttt{strat} planning call used temperature $0.0$ and a maximum output-token ceiling of 1600. These ceilings were not target lengths; task prompts constrained the requested response format directly.

\subsection{Quality Proxy Details}
\label{app:quality-proxies}

We pair established task-specific scorers where available with a deterministic slogan metric tailored to short-form template reuse: CLAUS for alternative uses, MAoSS for stories, and phrase-level non-template scoring for slogans.

\paragraph{Within-task standardization.}
For output $x$ in task $t$, let $Q(x)$ be the raw quality proxy. We use
\[
Q^z(x)=\frac{Q(x)-\mu_t}{\sigma_t},
\]
where $\mu_t$ and $\sigma_t$ are computed over all evaluated outputs for task $t$. For cell $c$ with $n=150$ outputs,
\[
Q^z_c=\frac{1}{n}\sum_{i=1}^{n}Q^z(x_{c,i}).
\]

\paragraph{AUT and story scores.}
For AUT, the CLAUS item is the target object and the response is the generated alternative use. For stories, the MAoSS item is the story prompt and the response is the generated story. Returned prediction scores are used as raw quality values before within-task standardization.

\paragraph{Slogan phrase-level non-template score.}
For slogans, the primary score measures phrase-template reuse within each slogan task. We normalize each slogan by lowercasing, standardizing quotation marks and apostrophes, stripping surrounding quotes, removing punctuation except word-internal apostrophes and hyphens, collapsing whitespace, and tokenizing words. We then extract all word bigrams and trigrams.

For each slogan task $t$, we count all bigrams and trigrams in the task-level slogan corpus. For slogan $x$, commonness is computed leave-one-out: the $n$-grams contributed by $x$ are subtracted before estimating how often its own phrases occur in the remaining corpus for the same task. Let $\overline{F}_{2,\mathrm{LOO}}(x)$ and $\overline{F}_{3,\mathrm{LOO}}(x)$ denote the mean leave-one-out commonness of the bigrams and trigrams in $x$. The phrase-boilerplate score is
\[
B(x)
=
0.45\,\overline{F}_{2,\mathrm{LOO}}(x)
+
0.55\,\overline{F}_{3,\mathrm{LOO}}(x).
\]
The raw slogan quality score is the negative within-task standardized boilerplate score:
\[
Q_{\mathrm{slogan}}(x)
=
-
\frac{B(x)-\mu_{B,t}}{\sigma_{B,t}}.
\]
Higher values therefore indicate less reuse of repeated phrase templates within the same slogan task. The metric uses only bigrams and trigrams; it does not use hand-curated slogan templates or unigram exclusion lists.

\paragraph{LLM-judge slogan robustness.}
As a robustness check, we replace the lexical slogan score with an LLM-as-a-judge creativity score. Each slogan is scored independently by GPT-5.4 on a 1--5 scale. The judge receives only the task context and one candidate slogan, not the provider, generation method, strategy, or treatment label.

The system prompt is:

\begin{quote}\small
You are an expert evaluator of short advertising and public-campaign slogans.

Your task is to score one candidate slogan for slogan creativity.

You will be given:
1. The slogan-writing task context.
2. One candidate slogan.

Do not infer or consider which model, method, prompt condition, or experimental treatment produced the slogan. That information is intentionally hidden.

Before answering, silently consider the slogan's task fit, clarity, fluency, slogan-likeness, memorability, persuasive appeal, and creative expression. Do not output your reasoning.

Score the slogan on an absolute 1--5 scale for slogan creativity.

For this task, slogan creativity means the slogan is both:
1. appropriate and effective for the specified product or campaign; and
2. creatively expressed as a concise slogan.

A high-scoring slogan should fit the task, be fluent and clear, feel slogan-like, and have memorable, distinctive, clever, or persuasive phrasing.

Do not reward novelty by itself. A strange, confusing, awkward, off-task, or unpersuasive slogan should receive a low score even if it is unusual. A fluent but generic slogan should receive a moderate score unless it is especially effective.

Use this scale:
1 = Poor: off-task, confusing, awkward, inappropriate, or not usable as a slogan.
2 = Weak: relevant but generic, flat, vague, cliched, or weakly persuasive.
3 = Adequate: relevant, fluent, and usable as a slogan, but not especially memorable or creative.
4 = Good: relevant, fluent, slogan-like, and clearly memorable, distinctive, clever, or persuasive.
5 = Excellent: highly effective and creatively strong; polished, memorable, persuasive, and immediately usable.

Do not penalize the slogan for exceeding the requested word limit. Judge the slogan's creative effectiveness as written.

Return only one integer from 1 to 5. Do not return JSON, explanation, punctuation, or any other text.
\end{quote}

The user prompt is:

\begin{quote}\small
Slogan-writing task context:

\emph{[task context]}

Candidate slogan:

\emph{[candidate slogan]}

Silently apply the rubric, then score this slogan for slogan creativity.

Return only one integer from 1 to 5.
\end{quote}

We score all 16,200 slogan outputs. The raw 1--5 scores are then standardized within the slogan task and substituted for the lexical slogan score in the full quality pipeline.

\paragraph{No-slogan robustness.}
We also rerun all quality-bearing analyses after dropping slogans entirely. In this version, quality is computed only from CLAUS-scored AUT responses and MAoSS-scored stories. 

\section{Supplementary Notes}\label{supplementary_notes}
\subsection{Technological Choices}\label{app:sentence-embeddings}

 Our main results are reported using \texttt{sentence-transformers/all-mpnet-base-v2} embeddings~\citep{reimers2019sentencebert}. We found the results robust under \texttt{bge-large-en-v1.5}~\citep{bge_embedding} and \texttt{e5-large-v2}~\citep{wang2022text} embedding models. These models are freely available on Huggingface and have been widely used in recent technological developments.

\subsection{Semantic-region construction for entropy}
\label{app:semantic-region-construction}

For \(D_{\mathrm{ent}}\), semantic regions were constructed at the prompt level rather than separately within each analysis cell. For each prompt condition \(p\), we pooled all generated outputs for that prompt across providers, generation methods, and instruction strategies. We L2-normalized the embeddings described in Section~\ref{app:sentence-embeddings} and fit K-means separately within each prompt condition. The number of clusters was fixed by task family: \(K_p=12\) for slogans, \(K_p=15\) for alternative uses, and \(K_p=12\) for stories. K-means used a random seed of 20260523 and 20 random initializations.

The \(K_p\) values were chosen heuristically to provide a coarse, fixed-resolution partition of each prompt's embedding space. We do not interpret them as estimates of the true number of semantic categories. Each output inherited its prompt-level cluster label, and each candidate pool's entropy was computed from its distribution over these fixed regions using the definition in Section~\ref{diversity_metrics}. Empty regions contributed zero to the entropy sum. The fixed denominator \(\log K_p\) makes entropy comparable across cells within the same prompt condition and rewards both broader region coverage and more even allocation across occupied regions. Rarefaction analyses used the same prompt-level region labels and denominator for all subpools.

\subsection{AI Usage}
We used Grammarly AI to improve the grammatical accuracy of the manuscript, and ChatGPT to speed up the implementation of standard statistical analysis code.

\section{Supplementary Tables}\label{app:tables}

\begin{table}[h]
\centering
\small
\setlength{\tabcolsep}{5.2pt}
\begin{tabular}{llcc}
\toprule
Method & Inst. &
$D_{\texttt{pair}}$ AUC &
$D_{\texttt{ent}}$ AUC \\
\midrule
\texttt{indep} & N &
0.36 [0.22, 0.54] &
0.40 [0.24, 0.59] \\
\texttt{indep} & D &
0.41 [0.23, 0.59] &
0.48 [0.19, 0.65] \\

\texttt{strat} & N &
0.51 [0.31, 0.69] &
0.57 [0.44, 0.70] \\
\texttt{strat} & D &
\textbf{0.55 [0.33, 0.72]} &
0.61 [0.54, 0.69] \\

\texttt{repr} & N &
0.47 [0.30, 0.63] &
0.52 [0.42, 0.69] \\
\texttt{repr} & D &
0.46 [0.31, 0.62] &
0.49 [0.34, 0.66] \\

\texttt{self} & N &
0.48 [0.30, 0.63] &
0.62 [0.52, 0.73] \\
\texttt{self} & D &
0.51 [0.32, 0.67] &
0.63 [0.50, 0.74] \\

\texttt{peer1} & N &
0.48 [0.31, 0.64] &
0.63 [0.53, 0.73] \\
\texttt{peer1} & D &
0.52 [0.33, 0.67] &
\textbf{0.64 [0.47, 0.74]} \\

\texttt{peer2} & N &
0.49 [0.31, 0.64] &
\textbf{0.64 [0.50, 0.73]} \\
\texttt{peer2} & D &
0.53 [0.33, 0.69] &
\textbf{0.64 [0.46, 0.76]} \\
\bottomrule
\end{tabular}
\caption{Rarefaction AUC summary for all generation settings. Values report mean AUC with 95\% percentile intervals in brackets. Higher values indicate faster accumulation across sampled subpool sizes. Bold marks the best displayed mean in each column. N denotes \texttt{neutral}; D denotes \texttt{diverge}.}
\label{tab:appendix-rarefaction-auc}
\end{table}

\begin{table}[h]
\centering
\small
\setlength{\tabcolsep}{5.2pt}
\begin{tabular}{llcc}
\toprule
Method & Inst. &
$D_{\texttt{pair}}$ target &
$D_{\texttt{ent}}$ target \\
\midrule
\texttt{indep} & N &
11.1 [2.0, 109.0] &
72.6 [6.0, 150.0] \\
\texttt{indep} & D &
3.6 [2.0, 17.0] &
17.9 [3.0, 69.0]\textsuperscript{a} \\

\texttt{strat} & N &
2.1 [2.0, 3.0] &
8.3 [3.0, 22.0] \\
\texttt{strat} & D &
\textbf{2.0 [2.0, 3.0]} &
7.4 [3.0, 21.0] \\

\texttt{repr} & N &
2.3 [2.0, 5.0] &
19.2 [3.0, 117.5] \\
\texttt{repr} & D &
2.4 [2.0, 5.0] &
23.6 [3.0, 111.0] \\

\texttt{self} & N &
2.1 [2.0, 4.0] &
7.3 [3.0, 19.0] \\
\texttt{self} & D &
2.1 [2.0, 3.0] &
\textbf{6.6 [3.0, 16.0]} \\

\texttt{peer1} & N &
2.1 [2.0, 3.0] &
7.1 [3.0, 18.0] \\
\texttt{peer1} & D &
\textbf{2.0 [2.0, 3.0]} &
6.9 [3.0, 16.0] \\

\texttt{peer2} & N &
2.1 [2.0, 3.0] &
7.0 [3.0, 16.0] \\
\texttt{peer2} & D &
\textbf{2.0 [2.0, 3.0]} &
7.2 [3.0, 18.0] \\
\bottomrule
\end{tabular}
\caption{First-hit rarefaction summary. Each entry reports the mean subpool size $q$ needed to match the corresponding metric value of the full 150-output \texttt{indep}--N pool, with 95\% percentile intervals in brackets. Lower values indicate faster accumulation; bold marks the best displayed mean in each column.}
\label{tab:appendix-first-hit-rarefaction}

\vspace{0.25em}
\raggedright
\footnotesize
\textit{Notes.} N denotes \texttt{neutral}; D denotes \texttt{diverge}. Targets are computed within the same bootstrap replicate, provider, and task family before aggregation. \textsuperscript{a}Target not reached in 525 of 2700 provider--task-family bootstrap curves.
\end{table}

\clearpage
\onecolumn

\section{Supplementary Figures}
\label{app:supplementary-figures}

\begin{figure*}[!ht]
\centering

\begin{subfigure}[t]{\textwidth}
    \centering
    \includegraphics[
        width=\textwidth,
        height=0.43\textheight,
        keepaspectratio
    ]{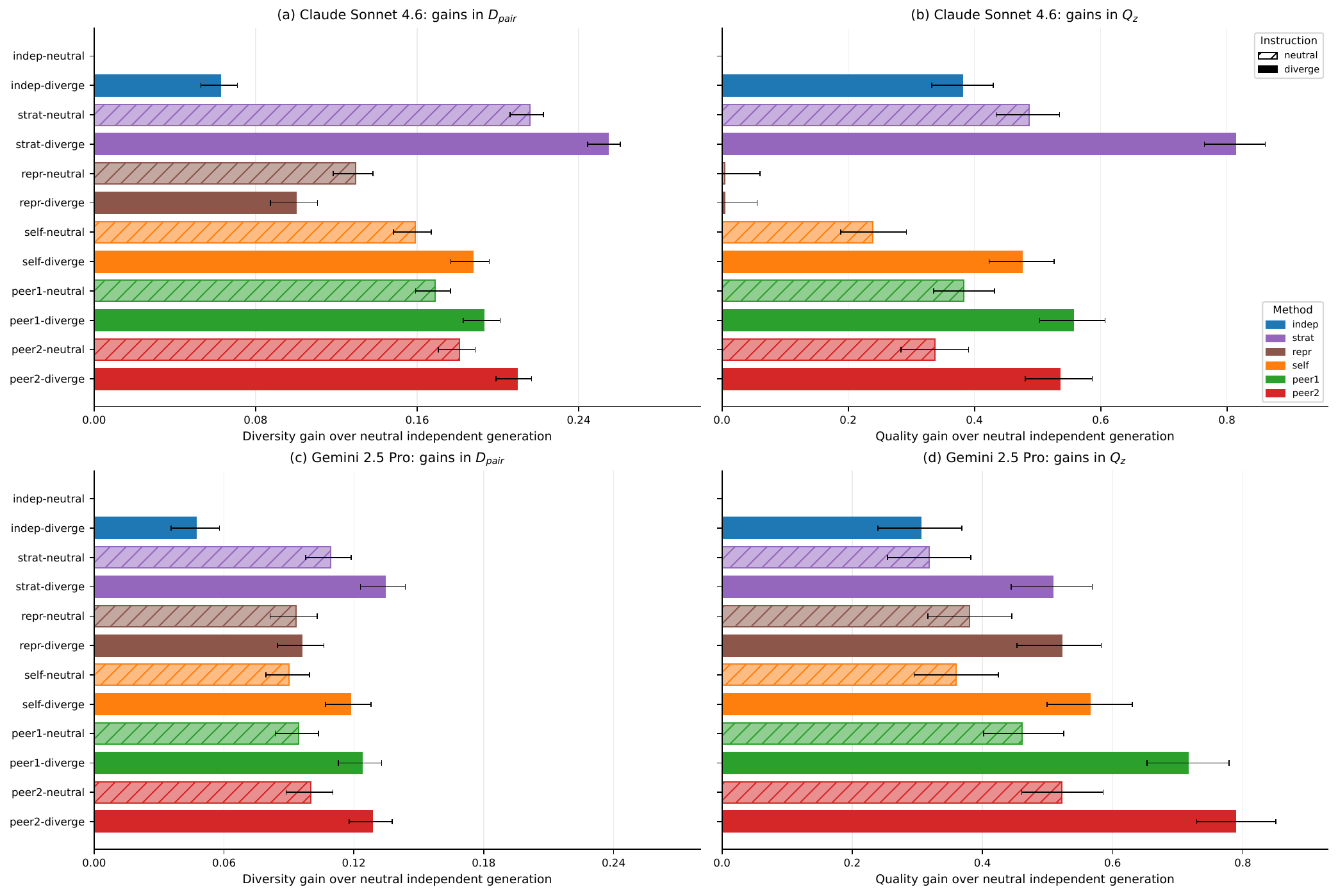}
    \caption{Provider-specific diversity and quality gains for Claude Sonnet 4.6 and Gemini 2.5 Pro. Bars report $\Delta^{\mathrm{base}}$ contrasts relative to \texttt{indep}--\texttt{neutral} within the same model and prompt condition; intervals show 95\% bootstrap intervals.}
    \label{fig:appendix-provider-dpair-quality}
\end{subfigure}

\vspace{0.6em}

\begin{subfigure}[t]{\textwidth}
    \centering
    \includegraphics[
        width=\textwidth,
        height=0.43\textheight,
        keepaspectratio
    ]{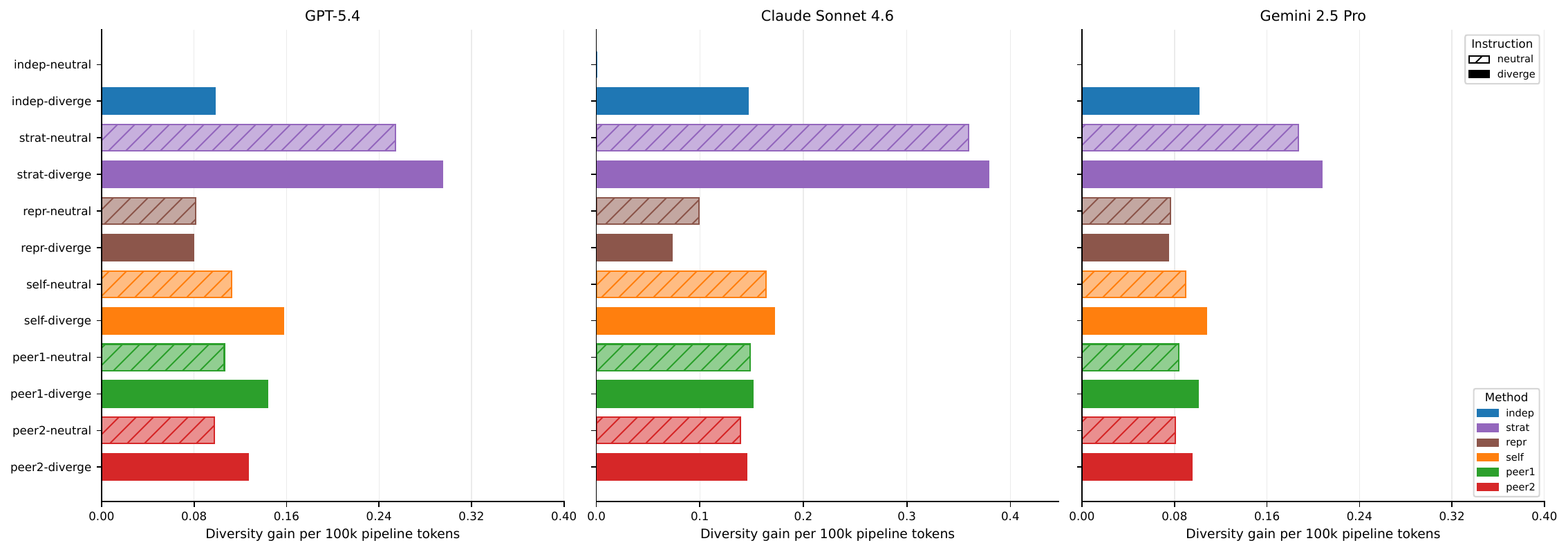}
    \caption{Token-normalized diversity gains for all providers. Bars report $D_{\texttt{pair}}$ gain per 100k full-pipeline tokens, using the same baseline contrasts as in Figure~\ref{fig:gpt54-frontier}.}
    \label{fig:appendix-token-efficiency-all-providers}
\end{subfigure}

\caption{Supplementary provider-level figures. Panel (a) reports provider-specific diversity and quality gains for the non-GPT models; panel (b) reports token-normalized diversity efficiency for all three providers.}
\label{fig:appendix-provider-and-token-summary}
\end{figure*}

\end{document}